\documentclass[letterpaper, 10 pt, conference]{ieeeconf}  

\IEEEoverridecommandlockouts                              

\usepackage[left=1.88cm,right=1.88cm,top=1.88cm,bottom=1.88cm]{geometry}
\usepackage{amsmath,amssymb,amsfonts}
\newtheorem{theorem}{Theorem}
\newtheorem{assumption}{Assumption}
\newtheorem{remark}{Remark}
\newtheorem{lemma}{Lemma}
\usepackage{graphicx}
\usepackage{textcomp}
\usepackage{xcolor}
\usepackage{bbm}
\let\labelindent\relax
\usepackage{enumitem}
\usepackage{tabularx, booktabs}
\usepackage[style=ieee, maxcitenames=2, mincitenames=1, natbib=true, dashed=false]{biblatex}
\usepackage{soul,color}
\usepackage{siunitx}
\usepackage{dsfont}

\usepackage{times}
\usepackage{multicol}
\usepackage{hyperref}
\usepackage{cleveref}
\usepackage{graphicx}
\usepackage{bm}
\usepackage[nolist]{acronym}
\usepackage[noend]{algpseudocode}
\usepackage[linesnumbered,ruled]{algorithm2e}

\usepackage{mathtools}
\usepackage{array}
\usepackage{dblfloatfix}
\usepackage{subfigure}
\usepackage{diagbox}
\usepackage{etoolbox}\AtBeginEnvironment{algorithmic}{\small}
\usepackage{color,soul} 

\usepackage{lipsum}
\usepackage{color}
\usepackage[colorinlistoftodos]{todonotes}
\newcommand{\blue}[1]{\protect\color{blue}{#1}\protect\color{black}}

\newcommand{\ego}{\operatornamewithlimits{e}}
\newcommand{\agent}{\operatornamewithlimits{o}}

\begin{document}

\title{
    Delayed-Decision Motion Planning in the Presence of Multiple Predictions
    
\author{
  David Isele, 
  Alexandre Miranda A$\tilde{\text{n}}$on,
  Faizan M. Tariq,
  Goro Yeh,
  Avinash Singh, and
  Sangjae Bae 
}
   
    \thanks{D. Isele, F.Tariq,G. Yeh, A.Singh, and S. Bae are with the Honda Research Institute, USA. {\tt\small $\lbrace$disele, faizan\_tariq, zheng-hang\_yeh,avinash\_singh,sbae$\rbrace$@honda-ri.com}. A. A$\tilde{\text{n}}$on conducted work while at Honda Research Institute, USA {\tt\small alex.miranyon@gmail.com} }
}
\maketitle

\begin{abstract}
Reliable automated driving technology is challenged by various sources of uncertainties, in particular, behavioral uncertainties of traffic agents. It is common for traffic agents to have intentions that are unknown to others, leaving an automated driving car to reason over multiple possible behaviors. This paper formalizes a behavior planning scheme in the presence of multiple possible futures with corresponding probabilities. We present a maximum entropy formulation and show how, under certain assumptions, this allows delayed decision-making to improve safety. The general formulation is then turned into a model predictive control formulation, which is solved as a quadratic program or a set of quadratic programs. We discuss implementation details for improving computation and verify operation in simulation and on a mobile robot. 
\end{abstract}

\section{Introduction}
	
Prediction technology continues to advance, and multiple prediction outputs are now a staple of state-of-the-art prediction methods \cite{tang2019multiple,choi2019drogon,phan2020covernet,salzmann2020trajectron++,shi2022motion,isele2024gaussian}. This paper examines how an autonomous driving (AD) agent can utilize multiple predictions in the behavior planning process. In the context of this work, behavior planning corresponds to the combined task of decision-making and trajectory planning \cite{slas}.

\begin{figure}
\centering
\includegraphics[trim={0cm 3cm 0 2.75cm},clip,width=0.65\columnwidth]{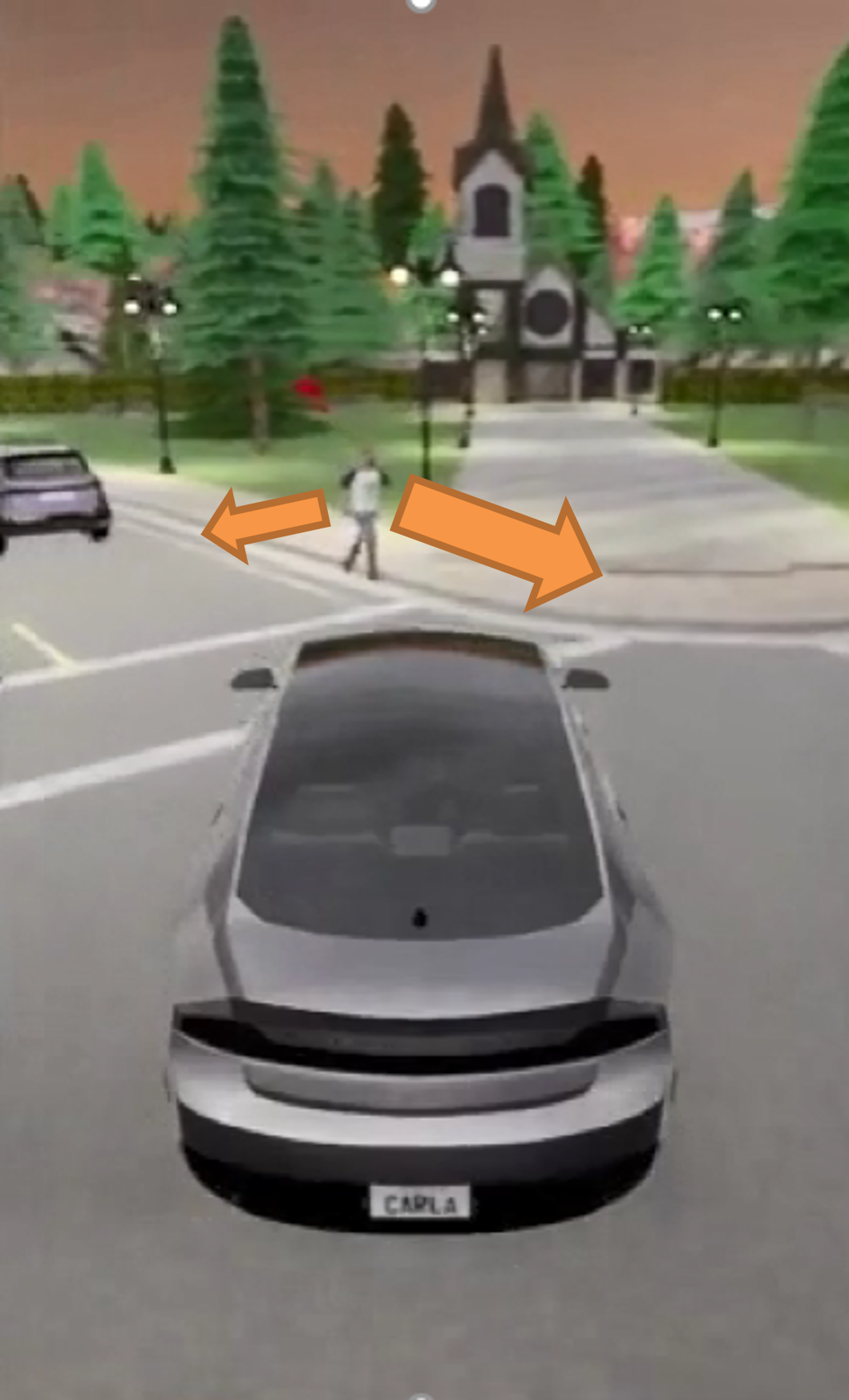}
\caption{\textbf{Illustrative example:} A pedestrian at the intersection has possible intentions of walking straight or turning right. The ego vehicle (at the center of the image) must comprehend that the pedestrian's potential behaviors reflect that planning, e.g., by slowing down enough to stop safely if needed. }
\label{fig:motivation}
\end{figure}

Consider the scenario depicted in Fig.~\ref{fig:motivation}. A pedestrian walks along a road and will likely continue straight (with 80\% probability), but the pedestrian is positioned close to the street, indicating that they \emph{might} turn to cross the street (with 20\% probability). Selecting the most probable sequence of events results in an overly aggressive and risky behavior; we assume they will not cross and are wrong 20\% of the time. However, a cautious policy unnecessarily brakes in the middle of an intersection for a pedestrian, creating confusing and potentially dangerous consequences. Intuitively, the solution in this instance is to try to balance the two behaviors: slow down so that it is possible (i) to brake if necessary, but (ii) not to make the overly conservative assumption that an unlikely event is true. This paper presents a formalism through which the problem of planning in the presence of multiple predictions can be expressed, where the intuitive solution of our example emerges as a result.

Starting with probabilistic predictions, we can adopt a behavior that maximizes the expected reward \cite{schmerling2018multimodal}. However, this is often insufficient for safety-critical applications because negative events, such as collisions, are discouraged but not forbidden. This shortcoming can be addressed by directly incorporating constraints into the formulation \cite{ahn2021safe}. However, such an approach is still overly conservative; even though marginalizing over an agent's distribution accurately assigns risk to both possible futures, it neglects that only one future can be true, generally resulting in a behavior that avoids both possibilities. 

One way around this difficulty is by delaying an agent's decision while selecting an action that keeps a response to both futures possible \cite{alsterda2019contingency,cui2021lookout}. In this work, we formalize the importance of delayed decision making and identify the optimal time to wait. 
 While the generalized framework for behaving under the uncertainty associated with multiple predictions can be carried out by various planning methods, to ground the formalism, we specifically detail its application to an MPC framework.
 The special case of solving the problem with MPC has some similarity to existing work \cite{alsterda2019contingency,chen2022interactive}. 
 Our approach solves parallel optimizations which was also explored by Adajania et al. \cite{adajania2022multi}, though in a different problem setting.
 Similar to our approach, Contingency MPC \cite{alsterda2019contingency} also solves for parallel plans with different costs and constraints, sharing a common initial state and action. However, they focus on steering under different friction assumptions, rather than handling predictions, and they issue a command for just a single time step. This avoids the decision process and the possibility of infeasibility. Branch MPC \cite{chen2022interactive} takes into account another agent's actions with corresponding probabilities, 
 but unlike our work, Branch MPC reasons over the other agent's entire action space which is substantially more computationally prohibitive, and therefore their investigation was limited to a single other agent.   
 
For an autonomous driving use case \cite{tariq2022autonomous}, there is a high non-convexity not addressed in many of the other works, which we show cannot be avoided in the worst case. Hence, we present reasonable approximations that allow us to handle the non-convexities efficiently. 
As part of the approximate solution, we develop a very fast piece-wise-linear trajectory planning approximation which can enable the quick evaluation of candidate solutions, and might be useful outside of our specific setting. Finally, to demonstrate the practicality of our algorithm, as well as the realtime computation speed, we showcase our results in simulation and on 1/10 scale cars.


\section{Problem Statement} \label{section:problem}
\subsection{Notation} \label{section:notation}
In this work, we focus on planning for a single agent of interest, referred to as the ego agent, and any variable $\psi$ pertaining to this ego agent is notated as $\psi^e$. Moreover, 
when it is clear from context that the ego agent is the one considered, the notation will be omitted.

As there will be a number of variables introduced in this manuscript, we list some of the important ones here to provide readers with a comprehensive reference.
\begin{itemize}
    \item $f_*$: True future
    \item $F$: Set of $m$ possible futures
    \item $c_*$: Set of spatio-temporal constraints pertaining to $f_*$
    \item $C$: Set of spatio-temporal constraints pertaining to $F$
    \item $\mathcal{T}_*$: Set of all feasible trajectories that satisfy $c_*$
    \item $t_R$: Time when the true future is revealed
    \item $t_d^{\ego}$: Last time instant where a feasible trajectory exists that satisfies all constraints in $C$ for the ego vehicle. 
\end{itemize}

\subsection{Problem Formulation}
The multi-future trajectory planning (MFTP) problem is posed as a stochastic game described by the tuple $\{\mathcal{S}, \mathcal{A}, P, \mathcal{R}\}$, where $\mathcal{S}$ is the world state which consists of a set of states $\{S^1,\dots,S^n\}$ for $n$ agents including the ego agent, $\mathcal{A}=\{A^1,\dots, A^n\}$ is the set of actions for agents, $P:\mathcal{S}\times \mathcal{A} \times \mathcal{S} \rightarrow [0,1]$ is the set of transition probabilities, and $\mathcal{R}=\{R^1,\dots, R^n\}$ is the set of rewards for each agent. The sets of feasible states, actions, and reward function~\eqref{eqn:reward_ego} for the ego agent are represented by $S^e$, $A^e$, and $R^e$, respectively. The ego agent generates a trajectory $\tau = \{s_1,\dots,s_T\}$ consisting of a set of states where $s_t \in S^e$ and $T$ is the time horizon.
A future $f = \{s_1^j,\dots,s_T^j\}, ~\forall j \in [1, \dots, n] \backslash{e}$, is defined as the set of future states for all agents starting from the current time. This implies that the behavior of each non-ego agent is fixed over a future $f$.

\begin{assumption} \label{assm_inclusiveness}
    \emph{Inclusiveness.} The true (unknown) future $f_* \in F$ is one of a known set of $m$ possible futures $F = \{f_1,\dots,f_m\}$.
\end{assumption}


In practice, Assumption \ref{assm_inclusiveness} allows us to make use of multi-intention prediction algorithms where the outputs can be taken directly from any off-the-shelf algorithm since there is no requirement that the predictions adhere to any particular constraints or have any particular distribution. Additionally, because we're considering the trajectory as a whole, we remove the large computational burden of Branch MPC \cite{chen2022interactive} since we no longer need to account for different actions from different agents at every time step. 

Note that a future is a joint prediction of all agents in the scene. 
To focus on the MFTP problem, we suppose Assumption~\ref{assm_inclusiveness} holds, such that one of the predictions is correct. Each future has a corresponding set of spatio-temporal
constraints $C=\{c_1,\dots,c_m\}$.


We overload the notation so that the reward of a trajectory conditioned on a future is given by 
\begin{equation}    \label{eqn:reward_ego}
\begin{aligned}
    R^e(\tau|f) &= \sum_{t=1}^T R^e(s_t^e,a_t^e|f) \\
    &= \sum_{t=1}^T R^e(s_t^1,\dots,s_t^n, a_t^1, \dots, a_t^n).
\end{aligned}
\end{equation}
Our objective is to maximize the ego agent's reward in Eqn.~\eqref{eqn:reward_ego}.
Given the true future $f_*$ and the reward function $R^e(\tau|f_*) \in \mathbb{R}$, our objective is to find an optimal trajectory $\tau_*\in \mathcal{T}_*$ that satisfies all constraints $c_*$ and maximizes the reward function $R^e$. Recall, $\mathcal{T}_*$ is defined to be the set of all feasible trajectories that satisfy $c_*$. 

Note that it is sufficient to satisfy the constraints pertaining to the true future, however, that is unknown. Therefore, for $i \in \{1,\dots,m\}$, we define the corresponding belief of future $f_i$ being $f_*$:  
\begin{equation}
p_i \coloneqq \mathds{P}(f_*=f_i).
\end{equation} 

\section{Approach}
We suppose that the probability of each future trajectory is invariant for each planning instance, which allows us to optimize the expected value:
\begin{eqnarray}\label{eq:expectation}
\max_\tau \mathbb{E}_{f\sim p(f)}[R(\tau|f)] = \max_\tau  \sum_{i=1}^m p_i R(\tau|f_i) \enspace .
\end{eqnarray}
However, in the case of safety critical systems, some trajectories will be catastrophic. We acknowledge this by allowing for some trajectories to have a corresponding reward of $-\infty$\footnote{in the case of autonomous driving, this will correspond to trajectories that result in collisions}. This makes it difficult to work with expectations, since \emph{all} expectations with non-zero probability of trajectories that result in $-\infty$ reward will have $-\infty$ expected value. We will therefore assert 
optimality can be sacrificed for the purpose of minimizing catastrophic events. We will refer to this assertion as the \textit{catastrophic assertion}.
\begin{assumption}\label{assm_catastrophe}
    \emph{Catastrophic Assertion.} Given a reward structure that allows $-\infty$ rewards, reducing the probability of $-\infty$ rewards is strictly more important than optimizations that make $\epsilon$-bounded improvements where $\epsilon$ is a finite value.
\end{assumption}

\begin{remark}
    Although the catastrophic assertion may not always hold true such that there could be instances where guessing a wrong future may not have $- \infty$ reward, we tackle the worst-case scenario for safety-critical systems, such as autonomous cars, in order to yield robust safety guarantees.
\end{remark}
\begin{remark}
    Utilizing the worst-case formulation, our algorithm places utmost importance on maintaining safety and predictability. However, in the event that safety cannot be maintained, i.e., incurring a $-\infty$ reward is inevitable, we can call upon our crash mitigation system to take evasive actions \cite{tariq2023risk}.
\end{remark}

\subsection{Maximum Probability Trajectory}
To ensure we have a viable action in response to an uncertain future, we want to maximize the probability that our trajectory is in the feasible set $\mathcal{T}_*$ of the true future $f_*$ at the time $t_R$ that the true future is revealed. We are therefore interested in maximizing the joint probability that we satisfy constraint $c_i$ and $\mathcal{T}_i$ is in $\mathcal{T}_*$ for each future $i$ at time $t = t_R$. A convenience of working with trajectories is that we know that if the ego agent is on $\tau$ at time $t_R$ and $\tau \in \mathcal{T}_*$, then it can stay on a feasible trajectory for all time after $t_R$. 

Let $\mathds{P}(s_t|s_{t-1},a_{t-1})$ denote the transition probability to $s_t$, 
and let $\mathbb{I}(s_t \models c_i)$ be an indicator function that is one if $s_t$ satisfies constraints $c_i$ and zero otherwise. Additionally, let the probability that the true trajectory is revealed at time $t$ be $\mathds{P}(t_R=t)$. 
Suppose transition probabilities are independent of future probabilities. 
The probability of having a feasible ego state (i.e., satisfying constraints) at the reveal time is then:
\begin{align}
    \mathds{P}(s_t \models c_*, t_R=t) &= \mathds{P}(s_t \models c_*) \cdot \mathds{P}(t_R=t) \nonumber \\
     = \sum_{i=1}^m p_i \mathds{P}(t_R=t) &\mathbb{I}(s_t \models c_i)\prod_{z=1}^t \mathds{P}(s_z|s_{z-1},a_{z-1}). \label{eq:step}
\end{align}
To find the probability that a particular $\tau$ is in $\mathcal{T}_*$ at the reveal time $t_R$ we can sum over all time steps. Under Assumption \ref{assm_inclusiveness}, we can simplify notation and return to reasoning in trajectory space as follows:
\begin{eqnarray}
    \mathds{P}(\tau \in \mathcal{T}_*) \coloneqq \sum_{i=1}^m p_i \sum_{t=1,\forall s_t \in \tau}^T \mathds{P}(t_R=t)\mathbb{I}(s_t \models c_i). \label{eq:maxprob}
\end{eqnarray}
In general, it is not possible to find one single trajectory that stays in all feasible spaces, as trajectories corresponding to different futures will inevitably diverge. However, \emph{we can} maximize the number of constraints we are able to satisfy by keeping multiple constraints satisfied for as long as possible. Some readers might correctly identify the connection to maximum entropy here, which we will discuss in section \ref{sec:maxent}. However, we will first discuss how keeping many trajectories feasible is equivalent to delaying a decision. 

\subsection{Delayed Decisions}

In this section we discuss how aligning with multiple solutions effectively delays our decision. We discuss why delaying can be beneficial, and explain how to identify the appropriate amount of time to delay. 

At some time $t=t_R$ we will know which future is the true future $f_*$. Trivially, this could be after time equal to the horizon time $T$ has passed, at which point the future will have already happened. In many cases, however, we can observe indications or definitive decisions in advance since certain actions make subsequent future actions impossible. Note that in order to make use of any observed indications or decisions, the ego agent needs sufficient time to react. 

If we know $f_i=f_*$, we can determine which states keep us in $\mathcal{T}_i$. Without knowing $f_*$ in advance, the last moment $t_d^{\ego}$ the ego agent can make a decision is defined as the point in time where there is at least one feasible trajectory that belongs to all trajectory sets $\mathcal{T}_1, \dots, \mathcal{T}_m$. Note that by symmetry, other agents have a similar $t_d^{\agent}$, after which point at least some futures will no longer be possible. 

If another traffic agent's $t_d^{\agent}$ occurs earlier in time than our agent's $t_d^{\ego}$, by staying in the overlapping space of feasible trajectory sets until time $t_d^{\agent}$ we can seize the opportunity and plan with reduced uncertainty. If time $t_d^{\ego}$ occurs earlier, then before $t_d^{\ego}$ happens, the ego agent should make a decision based on the current information.  

Note that if, as our policy, we choose to make a conservative decision immediately before $t_d^{\ego}$, we get a risk-free reduction in uncertainty (and increase in safety) for all cases where $t_d^{\agent}$ happens first. Additionally, 
if $t_d^{\ego}$ happens earlier, we can still benefit from waiting as a result of additional acquired information. The next section describes how to determine when to act in the case that $t_d^{\ego}$ occurs before $t_d^{\agent}$. 


\subsection{Using Time as a Proxy for Exploration}

When considering a delayed decision we begin with the observation that beliefs of agent behavior are not fixed. Beliefs become more accurate based on observations of the agent behavior. Over time, certain trajectories can be entirely discarded based on reachability analysis relating to the physical limits of the system. 

\begin{assumption} \label{assm_monotonicity}
\emph{Monotonicity of agent estimation.} The accuracy in the ego agents' belief of another agent's behavior monotonically increases with time. 
\end{assumption}
We will assume accuracy increases monotonically with information gathered to rule out the impact of noise for our proof. For the purposes of discussing time as a proxy for exploration, we will also assume that information gain is passive.
\begin{assumption} \label{assm_passive}
\emph{Information Gain is Passive.} We assume the ego agent is a passive observer collecting information.  
\end{assumption}
Assumption \ref{assm_passive} is generally not the case; there is a large collection of research on interactive agents  \cite{sadigh2018planning,yang2018cm3,isele2019interactive,tian2021anytime,hu2023active}. But for the purposes of this section, we assume the ego vehicle's behavior does not effect information gain to simplify the discussion on the importance of time. 


Given the assumptions  \ref{assm_catastrophe}, \ref{assm_monotonicity} and \ref{assm_passive},
we will show that if $t_d^{\ego}$ occurs before $t_d^{\agent}$, then $t_d^{\ego}$ is the optimal time to make a decision under the catastrophic assertion. We will first analyze the case of two possible futures and then extend it to multiple possible futures. 

\begin{theorem}\label{two_traj}
Suppose Assumptions \ref{assm_monotonicity} and \ref{assm_passive} hold. 
A rational agent waiting until the last possible time to decide $t_d$ between two possible solution sets $\mathcal{T}_1$ and $\mathcal{T}_2$ can only reduce the chance of infinite possible loss, for an added cost that is strictly finite. 
\end{theorem}
As a sketch of the proof we enumerate all possible cases of a choice made with less information and the corresponding rewards when the choice is either correct or incorrect. The full proof is in the Appendix\footnote{Appendix at http://tiny.cc/ICRA2025} in Section \ref{sec:proof1}. By Assumption \ref{assm_catastrophe}, the best strategy delays a decision until $t_d$. 

There can be multiple decisions and therefore multiple decision points over a given trajectory. Earlier decisions must be made first and often influence the ability to make later decisions. The following lemma concerns the uniqueness of the state for a given decision point. 

\begin{lemma}\label{lemma:points}
Uniqueness of Decision Points.
The decision point that occurs first in time corresponds to a unique connected region of state space. 
\end{lemma}
\begin{proof}
Suppose not, 
assume there are multiple feasibly reachable decision points at different points in state space. Since we cannot move instantaneously from one region to another, there was a previous decision that determined which of these two decision points an agent arrived at. This is a contradiction since this is the decision point that occurs first.
\end{proof}

\begin{theorem}
Suppose Assumption~\ref{assm_catastrophe} and \ref{assm_monotonicity} hold. A rational agent waiting until the last possible time to decide $t_d$ between multiple solution sets $\mathcal{T}_i\;\forall i \in \{1,...,m\}$ can only reduce the chance of infinite possible loss, for an added cost that is strictly finite.
\end{theorem}
\begin{proof} 
From Lemma \ref{lemma:points}, there is a unique decision point. At the time, trajectory sets will split. 
These trajectory sets can be grouped, by common action, into meta-trajectory sets that split on the given decision. The proof of Theorem \ref{two_traj} concludes the proof.  
\end{proof}
Note that the best decision at the first decision point is not not necessarily the max. We show an example where the greedy solution is sub-optimal in the Appendix in Fig.~\ref{fig:greedy}.

\subsection{Maximum Entropy}\label{sec:maxent}
To reach our final formulation for delayed decision making in the presence of multiple possible futures, we return to Eqn.~\eqref{eq:expectation} and regularize it with the the maximum entropy formulation \cite{ziebart2008maximum,eysenbach2019if}
\begin{eqnarray}
\max_\tau \mathbb{E}_{f\sim \mathds{P}(f)}[R(\tau|f)] + H(\tau) \enspace .
\end{eqnarray}
Using the formulation for the maximum probability trajectory from 
Eqn.~\eqref{eq:step} and \eqref{eq:maxprob} this becomes
\begin{align}
    \max_\tau  \sum_{i=1}^m p_i R(\tau|f_i)
    & - \mathds{P}(\tau \in \mathcal{T}_i) 
    \log\big(\mathds{P}(\tau \in \mathcal{T}_i)\big) \enspace \\
    =    \max_\tau  \mathbb{E}_{f_i \sim \mathds{P}(f)}  &\bigg[ \nonumber
    R(\tau|f_i)  
    - \sum_{t=1,\forall s_t \in \tau}^T \mathds{P}(t_R=t)\mathbb{I}(s_t\models c_i) \\
    &\log\bigg(p_i
    \sum_{t=1,\forall s_t \in \tau}^T \mathds{P}(t_R=t)\mathbb{I}(s_t \models c_i)
    \bigg) \bigg] \enspace.\label{eq:ind}
\end{align}

In Eqn.~\eqref{eq:ind}, $p_i$ is fixed, so intuitively\footnote{
To arrive at the same result rigorously note that the second term has the form $xlog(ax) = xlog(a)+xlog(x)$. The maximum can then be confirmed to be at $x=1$, by checking the boundary conditions and critical point and noting the limit of xlog(x) is 0 as x approaches 0.
}, our only tool to effect the entropy is whether the indicator function is active, and having all indicators active will produce the maximum entropy. Having all indicators active will not, in general, result in the maximum reward. However, following the catastrophic assertion in Section \ref{section:problem}, we will assume any finite penalties suffered by the reward are negligible given it allows us to ensure a valid response for all possible futures up to $t_d$ seconds. The entropy is given priority by making it a hard constraint, and is solved by locking states up to $t_d$. This results in the formulation
\begin{align}\label{eq:general}
&\max_{\tau} \sum_{i=1}^m p_i R(\tau | f_i) \\
\mbox{s.t. } \tau \models \mathcal{T}_i, &\forall i \in \{1, \cdots, m\}, \forall t \in \{0,\dots,t_d\} \nonumber \enspace .
\end{align}

\section{Implementation}\label{sec:implementation}
Equation \eqref{eq:general} is a general formulation which can be realized through various methods including Model Predictive Control (MPC), Graph Search, and Reinforcement Learning. In this work we will focus on an MPC formulation.



For our MPC formulation, we decouple path planning and speed planning where the multiple predictions are handled by the speed planner. We convert the constraints associated with a prediction to piecewise linear upper and lower bounds. Constraints are typically not quadratic, so we generate multiple basins or ``profiles'' following the work by A$\tilde{\text{n}}$on et. al  \cite{anon2024multi}. The introduction of multiple futures increases the number of quadratic programming problems to be solved, which we address in Section \ref{sec:mp}.

We note that existing works have also addressed similar problems:
Contigency MPC solves for parallel plans \cite{alsterda2019contingency}, and Interactive multi-modal motion planning develops a branching MPC
\cite{chen2022interactive}. Unlike our work, these other works do not consider more than a single other agent.

\subsection{Quadratic Program Formulation}
Given the piecewise linear bounds obtained from a cell planner \cite{anon2024multi}, we simultaneously solve for multiple trajectories corresponding to multiple futures where all trajectories are locked up to time $t_d$. The decision time $t_d$ could be found exactly by binary search, however, to reduce computation, an acceptable look ahead time is used as a tunable parameter with a backup plan of the previous most probable solution in the event that the solver encounters an infeasibility. 


The MPC objective function is designed to promote comfort and reduce travel time, 
\begin{align}
    \min_x\;\; &\frac{1}{2}x^\top W x + x^\top q \\
    \mbox{s.t.  } &lb_1 < \mathcal{T}_1 < ub_1 \\
    &\dots \nonumber\\
&lb_m < \mathcal{T}_m < ub_m \\
&Zx = 0 
\end{align}
where $x$ is a concatenated position, velocity, acceleration, and jerk for each time step 
for each trajectory weighted by probability: 
\begin{eqnarray}
x = \langle\mathcal{T}^{0:t_d},p_1 \mathcal{T}_1^{t_d:T},\dots,p_m\mathcal{T}_m^{t_d:T}\rangle
\end{eqnarray}
$W \in \mathbb{R}^{\dim(x)\times \dim(x)}$ is a diagonal matrix that encodes the weights for smoothness, $q$ is zero everywhere except the final displacement which is used to encourage large displacement from the starting position thereby reducing travel time, the piecewise lower and upper bounds enforce the safety constraints, and $Z$ enforces the initial conditions and vehicle dynamics. 

\begin{figure*}[ht!]
\centering
\includegraphics[width=1.0\columnwidth]{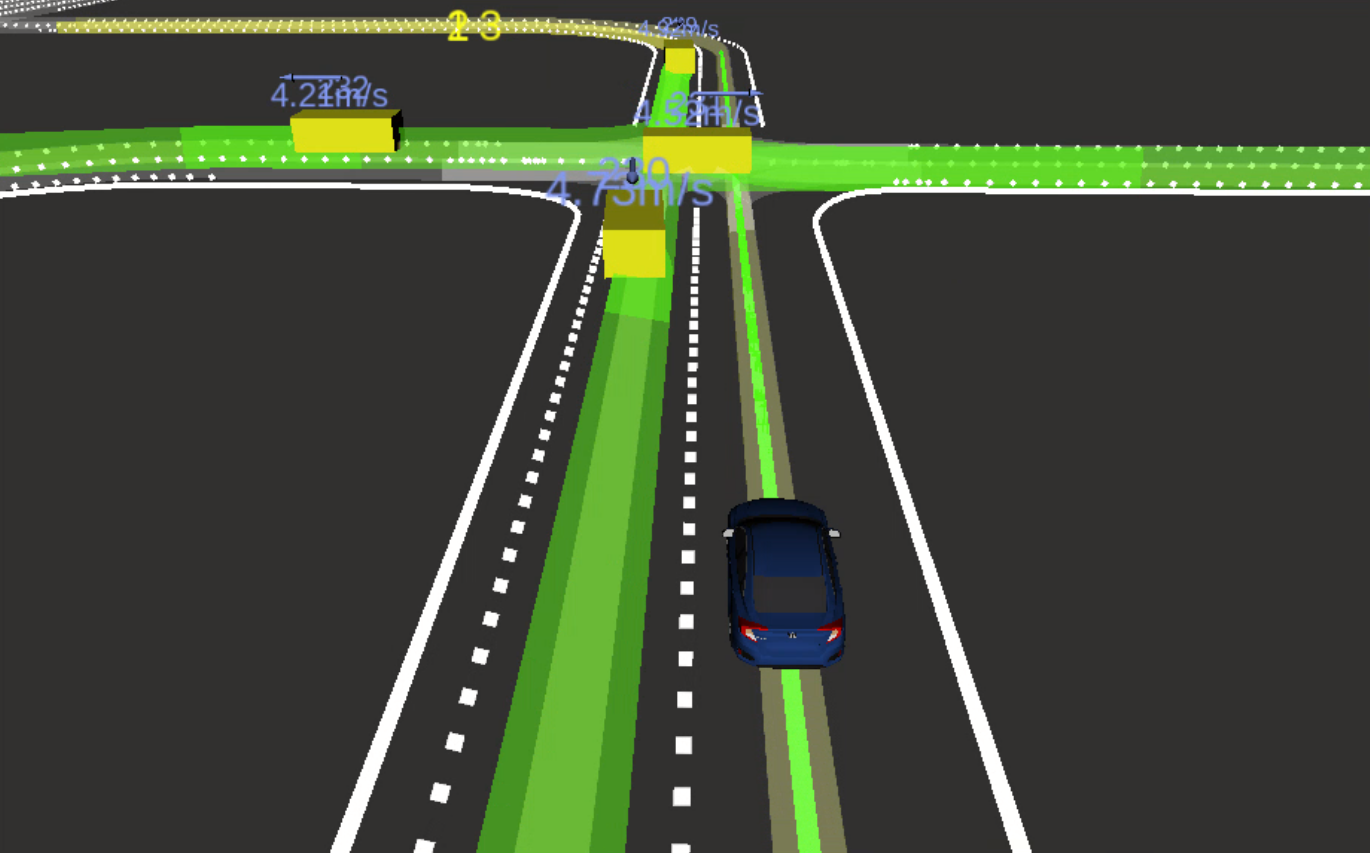}
\hspace{20pt}
\includegraphics[width=0.43\columnwidth]{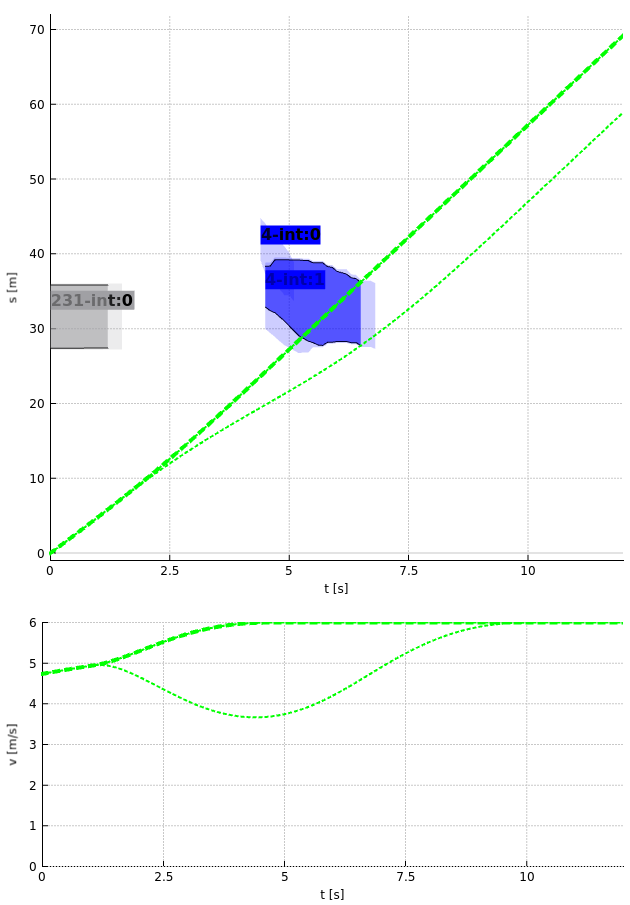}
\caption{Validation in CARLA: (Left) Depiction of the scene in RVIZ. (Right) Diagram of the space-time graph showing the planned trajectory and corresponding velocity. The ego vehicle approaches an intersection. The car currently crossing the intersection is depicted in gray on the ST graph. The car in the oncoming lane  (shown in blue on the ST graph) is predicted as potentially turning, and potentially going straight. At 1s, the trajectories are matched. After 1s, their are two trajectories corresponding to continuing straight and yielding for the turning car.}
\label{fig:carla}
\end{figure*} 
\subsection{Handling Complexity}\label{sec:mp}

The ST cell planner \cite{anon2024multi} quickly finds all sets of constraints where each set of constraints corresponds to a different quadratic program. However, if there are $k$ traffic agents, in the worst case, there may be $\bar{c} = 2^k$ sets of quadratic constraints (quadratic programming problems), see Fig.~\ref{combinatoric} in Section \ref{proofs} of the Appendix for an example of this worst case complexity. This complexity is compounded when considering multiple futures.

There are $m$ joint predictions of the future, however many prediction algorithms only output single agent predictions. If each car has $r$ predictions, there will be $m=r^k$ possible futures resulting from the different future combinations for each agent. Note, that if we were instead to consider the full action space of each agent, there would be an exponential number of possibilities being compounded at \emph{every} time step. Additionally, when passing constraints for each future to the multi-future optimization, the selection of constraints is combinatoric with $\bar{c}^m$ possible problems as you try to optimize against the different sets of constraints for each future. This results in an $O(2^{k^{r^k}})$ number of optimization problems in the worst case, which is clearly computationally intractable. Here we discuss approximations that can be used to reduce the number of problems to a manageable amount. Section \ref{sec:visualization_of_complexity} in the Appendix includes an example that helps illustrate the computational complexity of this section. 

\begin{figure}[ht!] 
\centering
\includegraphics[trim={0cm 0cm 0 1.3cm},clip,width=0.7\columnwidth]{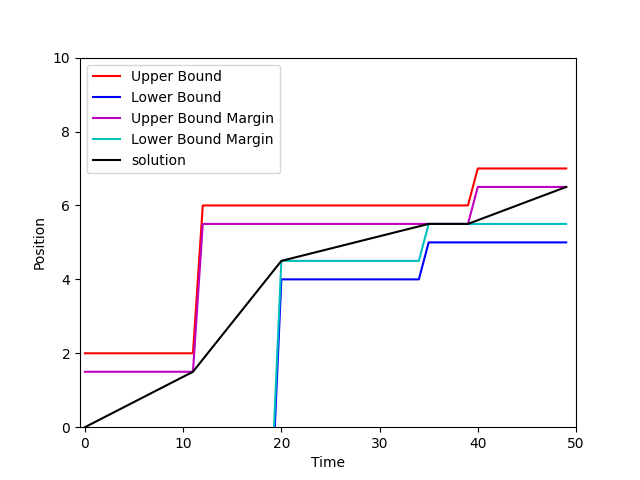}
\caption{Example visualization of the piece-wise linear trajectory planner generated by Algorithm \ref{alg:approx}.} \label{piecewise} 
\end{figure}
\subsubsection{The number of constraints}
While  
a combinatorial number of convex optimization problems can exist, in practice, the number of cars predicted to overlap the ego vehicle's trajectory is often a small number for short horizons (tens of seconds) for any given scene. In many of the cases when there are a large number of agents overlapping the ego car's path, such as congested highway merges, the number of problems scale linearly with $k$ since the cars don't cross each other's paths. So in practice,  $\bar{c} \approx \Theta(k)$ 
This may still be too large of a number of problems to solve quickly, so we look at a way to quickly analyze and discard infeasible and clearly sub optimal solutions. 
\begin{algorithm}[hbt!]
\small
\caption{Approximate profile from bounds}\label{alg:approx}
\textbf{Input: } lower bound $lb$ and upper bound vectors $ub$ where the length is prediction horizon T time step. Starting point $pt_0$ corresponding to the ego car's current position in the ST graph.  \\
\textbf{Output: }{piece-wise linear approximate solution $solution$}\\
$max\_margin \gets min((ub-lb)/2.0)$ \mbox{\blue{Find the maximum possible safety margin}} \\
pad margins with the $max\_margin$ \\
$pt_T \gets \mbox{ padded upper bound at time } T$ \mbox{\blue{ Note: we choose this end point to maximize travel distance}}
$solution \gets lower\_split(lb_{padded},ub_{padded},pt_0, pt_T,new\_split=True)$ \\
\textbf{return:} $solution$
\end{algorithm}
\begin{algorithm}[hbt!]
\small
\caption{Lower split}\label{alg:lower}
\textbf{Input: } lower bound $lb$ and upper bound vectors $ub$. Starting point $pt_0$, and ending point $pt_E$, flag $new\_split$ indicating this is the first call. \\
\textbf{Output: }{piece-wise linear approximate solution $solution$}\\
$proposal \gets linear\_interp(pt_0, pt_E)$ \\
$diff \gets proposal - lb$ \\
$violation \gets count(diff<0)$  \mbox{\blue{Negative values indicates the bound is being violated}} \\
\If{number of violations is 0}{
    \If{$new\_split == True$ }{
    \textbf{return:} $upper\_split(lb,ub,pt_0,pt_E, False)$
    }
    \Else{
    \textbf{return:} $proposal$
    }
}
$s\_id \gets argmin(diff)$ \mbox{\blue{Find the largest violation to use as the separation point}} \\
$first\_half \gets upper\_split(lb_{0:s\_id},ub_{0:s\_id},pt_0,lb[s\_id], True)$\\
$second\_half \gets upper\_split(lb_{s\_id:end},ub_{s\_id:end},lb[s\_id],pt_E, True)$\\
\textbf{return:} $concat(first\_half,second\_half)$
\end{algorithm}
\subsubsection{Quick evaluation of constraints}
To quickly filter out infeasible and clearly suboptimal problems, we develop a fast approximate solution. The fast approximate solution returns a piecewise-linear speed profile that maximizes the minimum margin to the bounds.
An example approximate solution is shown in Fig.~\ref{piecewise}. These approximate solutions let us quickly assess the feasibility of the trajectories given the constraints. Our C++ implementation runs in 2e-06 seconds on an Intel Core i9-10920X CPU. 

As a pre-processing step we turn the bounds into monotonic functions, which is equivalent to making the assumption that our speed planner cannot go in reverse. These bounds are then used as inputs to the \texttt{approximate profile from bounds} in Algorithm \ref{alg:approx}. The algorithm is a divide-and-conquer algorithm that splits the problem into sub-problems at the maximum violation of upper or lower bounds. 

The algorithm starts by identifying the maximum safety margin, and then pads the bounds by that amount. The algorithm then calls the lower split algorithm described in Algorithm \ref{alg:lower}. Algorithm \ref{alg:lower} and Algorithm \ref{alg:upper} (in the appendix) are alternating checks to verify if the problem or sub-problem satisfies the bounds. Algorithm \ref{alg:upper} closely mirrors Algorithm \ref{alg:lower}, and was moved to the appendix for space. If the bounds are violated, the problem is decomposed into sub-problems where the bounds are guaranteed to be valid. Upper split and lower split are called recursively until all sub-problems are valid, at which point the sub-problems are returned and concatenated into the full solution. Because both the upper split and lower split must be called at least once, there is a flag to ensure correctness. Proofs of the validity of the algorithm 
are included in Section \ref{alg_proofs} of the Appendix. 

\subsubsection{Pairing constraint sets}
To reduce the combinatorial effect between the number of futures and the number of constraints, we pair the constraint sets for each future. We pair constraints based on the similarity of approximate trajectories. This is reasonable as largely diverse trajectories often produce infeasible or highly suboptimal solutions, however, it is possible that the optimal solution is not the paired trajectory. 
The pairing heuristic reduces the $O(\bar{c}^m)$ complexity to $O(\bar{c})$, but we lose our ability to guarantee optimality. 

\subsubsection{Agent futures}
The complexity of working with multiple agents who can each select their actions independently is a common concern in game theory formulations. There is much effort to reduce the complexity for these problems, for example,  using learning based strategies \cite{foerster2018counterfactual,yang2018cm3,ma2022recursive} which can hopefully identify and exploit correlations in the data. Another common approach \cite{hu2023active} is to consider only a small number of agents of interest to have multiple futures
\cite{sadigh2018planning,tian2021anytime,chen2022interactive}. The QP problem has a sparse matrix that scales quadratically with the number of futures. 
\begin{figure}[ht!] 
\centering
\includegraphics[trim={2cm 0cm 0 0cm},clip,width=0.32\columnwidth]{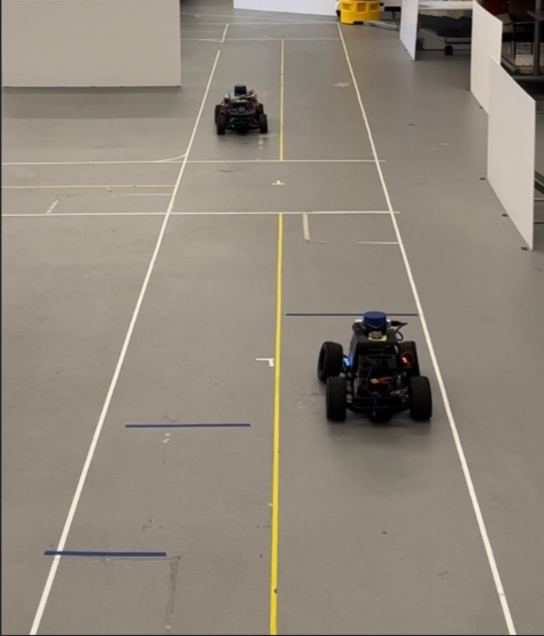}
\includegraphics[trim={0cm 0cm 0 0cm},clip,width=0.6\columnwidth]{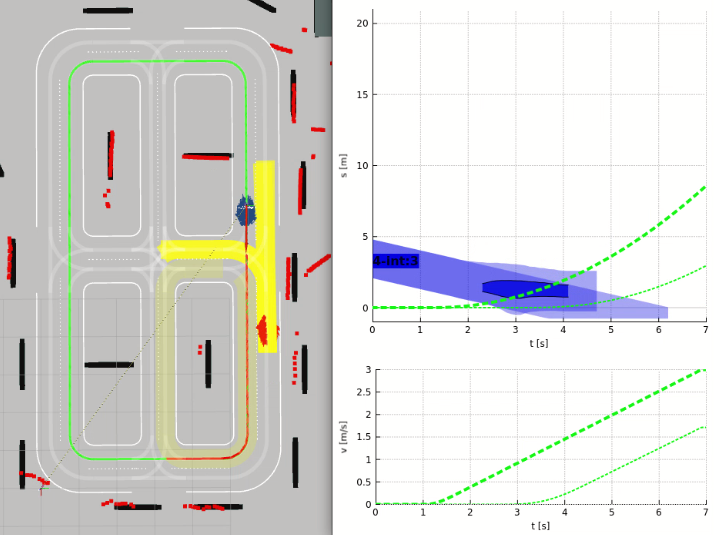}
\caption{Validation run on 1/10 scale RC cars. (Left) A snapshot of the scenario; (Center) An RVIZ visualization with predictions in yellow. (Right) A visualization of different plans corresponding to different predictions.} \label{fig:rc} 
\end{figure}

Summarizing the impact of these approximations, we start with $\bar{c}^m = O(2^{k^{r^k}})$ problems of size $O(r^k)$. Pairing constraint sets reduce this to $O(2^k)$ problems of size $O(r^k)$. In many real world driving scenarios, this is closer to $\Theta(k)$ problems of size $O(r^k)$. The problem size is then reduced to $O(r^l)$ where $l<4$ by restricting the number or agents with multiple futures. Some of these problems can be quickly discarded with a preliminary analysis made possible by Algorithm \ref{alg:approx}. We find that when $k<16$ and $m < 8$, the system can run in realtime. 



\section{Validation}\label{sec:result}
To validate our algorithm, we run our implementation in the simulator and on a 1/10 scale car. Traffic vehicles follow a constant velocity policy and GLK \cite{isele2024gaussian} is used for prediction. 
Figure \ref{fig:carla} shows an example of a scenario in CARLA. 
An MPC formulation that only predicts the car moving forward would be at risk if the car turns, or conversely, immediately slows down, taking the possibility of a turn as given. The multi-future planner is able to handle multiple predictions. We also observe that because the decision is delayed, we do not immediately brake for a turning car, but are able to slow down enough so that we can brake in the future if needed. Figure \ref{fig:rc} depicts a scenario with 1/10 scale RC. 
In the plan, the agent waits long enough to determine if the car is passing, where it will then accelerate or continue to wait accordingly. An MPC formulation with only the (most likely) forward prediction will immediately accelerate and then not be able to respond when the prediction is updated.














\printbibliography

\clearpage
\appendix
Here we include supplemental material to the main text.

\subsection{Example in ST graphs} 
\begin{figure} [ht]
\centering
\includegraphics[width=0.45\columnwidth]{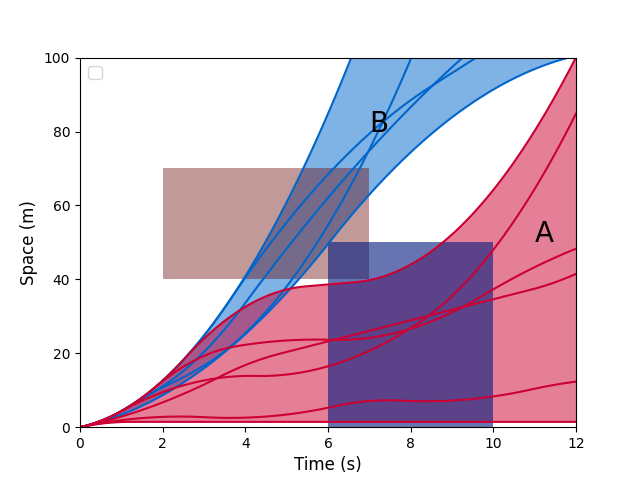}
\includegraphics[width=0.45\columnwidth]{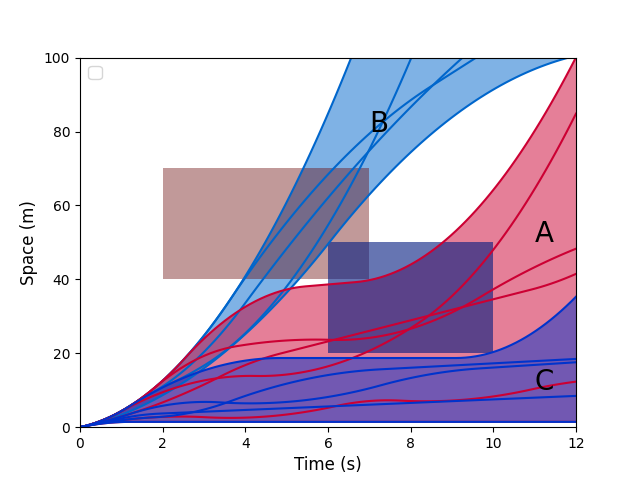}
\caption{Example ST graph.}
\label{fig:traj_example}
\end{figure}

This section provides a visual example of a motivating problem for multi-future planning using a space-time (ST) graph. The left diagram in Fig.~\ref{fig:traj_example} shows an ST graph which allows us to visualize an agents temporal traversal along a path. Obstacles that block the path are depicted as rectangles, where the different obstacle correspond to two different possible futures. Feasible trajectories corresponding to the red obstacle are depicted in region A. Feasible trajectories corresponding to the blue obstacle are depicted in region B. Given that beliefs are \textbf{fixed} and we \textbf{cannot influence} other agents, there is no possible trajectory that will satisfy both possible futures. So in this case, selecting the more probable future and optimizing the trajectory in that corresponding region is optimal, but has an expected reward of $-\infty$ since it will collide with a non-zero probability. 

The presentation of this example assumes we need to decide now. However if we delay our decision this can allow a better decision once more information is acquired. This is the motivating idea behind our work and raises the questions "When should we decide?" and "what do we do until we decide?" 

The right diagram in Fig.~\ref{fig:traj_example} depicts a similar situation, except now there is a safe trajectory that is more conservative, so if we were using the Expectation in Eqn.~\eqref{eq:expectation}, we would choose the conservative trajectory from region C. However, since we know that only one future is possible, we could stay in the overlap of A and B, and make a decision at 4s, at which point we should know whether A or B corresponds to the true future. 

\subsection{Proof of Theorem \ref{two_traj}}\label{sec:proof1}

\begin{proof}
Without loss of generality, we prove for an arbitrary $(f_1,f_2) \in F$ and show that Theorem~\ref{two_traj} holds at all possible cases. Suppose you make a decision at the last possible time $t_d$, your expected reward is 
\begin{align} 
\mathbb{E}_{f\sim \mathds{P}(f)}[R(\tau|f)] &= \nonumber \\ 
\mathds{P}(f_*=f_1)*R&(\tau_{t_d}|f_1) + \mathds{P}(f_*=f_2)*R(\tau_{t_d}|f_2) \enspace,
\end{align} 
where $\tau_{t_d}$ is the trajectory that results when making a decision at time $t_d$. 
Alternatively, if you make an earlier decision at alternate time $t_a$ (i.e., $t_a < t_d$), the expected reward is 
\begin{eqnarray} 
\mathds{P}(f_*=f_1)*R(\tau_{t_a}|f_1) + \mathds{P}(f_*=f_2)*R(\tau_{t_a}|f_2) \enspace.  
\end{eqnarray} 
Without loss of generality we will assume the true future is future 1. 

\textbf{Case 1: we compare the case of the agent choosing $\tau_1$ at time $t_d$ vs. the case of the agent choosing $\tau_1$ at time $t_a$.}
Because the agent is less constrained when deciding at time $t_a$, the trajectory can be optimized to a better but at least as good solution.
\begin{eqnarray} 
R(\tau_{t_d}|f_1) \le R(\tau_{t_a}|f_1)
\end{eqnarray}
However, since both agents chose trajectories corresponding to the true future, all $-\infty$ rewards are avoided
\begin{eqnarray} 
R(\tau_{t_d}|f_1) + k_1 = R(\tau_{t_a}|f_1)
\end{eqnarray}
for some finite $k_1$.

\textbf{Case 2: we compare the case of the agent choosing $\tau_2$ at time $t_d$ vs. the case of the agent cohosing $\tau_2$ at time $t_a$}
Here, both agents choose a catastrophic trajectory
\begin{eqnarray}
R(\tau_{t_d}|f_1) = R(\tau_{t_a}|f_1) = -\infty
\end{eqnarray}

\textbf{Case 3: we compare the case of the agent choosing $\tau_1$ at time $t_d$ vs. the case of the agent choosing $\tau_2$ at time $t_a$}
\begin{eqnarray} 
R(\tau_{t_d}|f_1) = k_2 \\
R(\tau_{t_a}|f_1) = -\infty
\end{eqnarray}
for some finite $k_2$.

\textbf{Case 4: we compare the case of the agent choosing $\tau_2$ at time $t_d$ vs. the case of the agent choosing $\tau_1$ at time $t_a$}. 
Given that accuracy in belief is strictly non-decreasing (Assumption~\ref{assm_monotonicity})
and the agents are rational (i.e., a trajectory with higher belief is selected), this case is not possible. 

In case 1, the agent has a potential loss of $k_1$, and in case 3, the agent has a potential gain of $k_2+\infty$. For an evaluation of the optimal reward, the likelihood of case 1 and case 3 would need to be determined, which will be scenario specific, but the analysis of the particular cases is sufficient to show that the possible benefit of an earlier decision is finite, where as the potential loss of an earlier decision is infinite.  
\end{proof}





\subsection{Worst Case Proof for Number of QPs}\label{proofs}
\begin{figure}[ht] 
\centering
\includegraphics[width=\columnwidth]{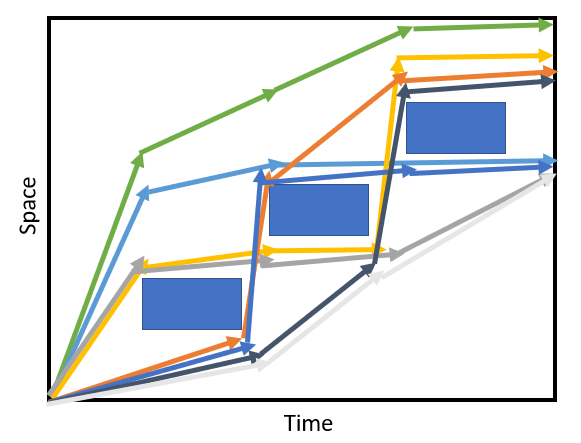}
\caption{QP problem growth. This contrived example shows the worst-case scaling. Going above or below an obstacle results in $c = 2^k$ possible quadratic problems for $k$ obstacles in the worst case.}\label{combinatoric}
\end{figure}

The first proof considers the number of quadratic problems associated with constraints imposed by multiple traffic agents. We visualize a traffic configuration in the ST graph in Fig.~\ref{combinatoric}. 
The colored arrows indicate basins corresponding to quadratic programming problems. This example shows a worst-case example showing how the number of constraints scales exponentially with the number of traffic participants. Note that this is a contrived example where the ego agent is driving through a scene similar to the frogger video game.  

\subsection{Approximate Speed Profile from Bounds}\label{approx_alg}
Here we provide the pseudo-code for the Approximate Speed Profile described in Section \ref{sec:implementation} and depicted in Figure \ref{piecewise}. The Approximate Speed Profile from Bounds Algorithm \ref{alg:approx} uses divide-and-conquer strategy with alternating calls to Algorithm \ref{alg:lower} and Algorithm \ref{alg:upper}.

\begin{algorithm}[hbt!]
\small
\caption{Upper split}\label{alg:upper}
\textbf{Input: } lower bound $lb$ and upper bound vectors $ub$. Starting point $pt_0$, and ending point $pt_E$. \\
\textbf{Output: }{piece-wise linear approximate solution $solution$}\\
$proposal \gets linear\_interp(pt_0, pt_E)$ \\
$diff \gets ub - proposal$ \\
$violation \gets count(diff<0)$  \mbox{\blue{Negative values indicates the bound is being violated}} \\
\If{number of violations is 0}{
    \If{$new\_split == True$ }{
    \textbf{return:} $lower\_split(lb,ub,pt_0,pt_E, False)$
    }
    \Else{
    \textbf{return:} $proposal$
    }
}
$s\_id \gets argmin(diff)$ \mbox{\blue{Find the largest violation to use as the separation point}} \\
$first\_half \gets lower\_split(lb_{0:s\_id},ub_{0:s\_id},pt_0,ub[s\_id], True)$\\
$second\_half \gets lower\_split(lb_{s\_id:end},ub_{s\_id:end},ub[s\_id],pt_E, True)$\\
\textbf{return:} $concat(first\_half,second\_half)$
\end{algorithm}

\subsection{Approximate Algorithm Proofs}\label{alg_proofs}

\textbf{Runtime} In the general, there can be at most $T$ splits. This can happen for instance in the case of an adversarial input where the upper and lower bounds create a narrow pass like between a staircase pattern. Since each split examines at most $T$ points when checking for the violation, the worst case runtime is $O(T^2)$. However, in an autonomous navigation setting, the steps result from the bounding box of another agent. This means that with $k$ agents there can be at most $O(k)$ changes in the bounds resulting in a runtime of $O(kT)$. In practice, this will be less as only agents that cross the ego agent's path are considered. 

\textbf{Proof of Validity}
We want to ensure the returned speed profile does not violate the bounds, this is done through the flags that ensure each segment gets verified by both upper and lower bounds. We'll use a proof by contradiction, suppose there is a violation in the upper bound along some line segment. We first assume that line segment was generated as the proposal in \emph{lower split}. \emph{Lower split} only returns the proposal if $new\_split$ flag is false, meaning \emph{upper split} generated the proposal, evaluated it, and found no violations. This is a contradiction. Now we assume that line segment was generated as the proposal in \emph{upper split}, but \emph{upper split} immediately checks proposals for violations, and only returns proposals without upper bound violations. So this is also a contradiction. By a symmetric argument we also find there was no violation along the lower bound. 





\subsection{Visualization of Computational Complexity}\label{sec:visualization_of_complexity}
This section provides visualizations to help make the various components of computational complexity clear, and also show what the numbers look like in a natural scene with our described approximations. 
\begin{figure} [ht]
\centering
\includegraphics[width=0.8\columnwidth]{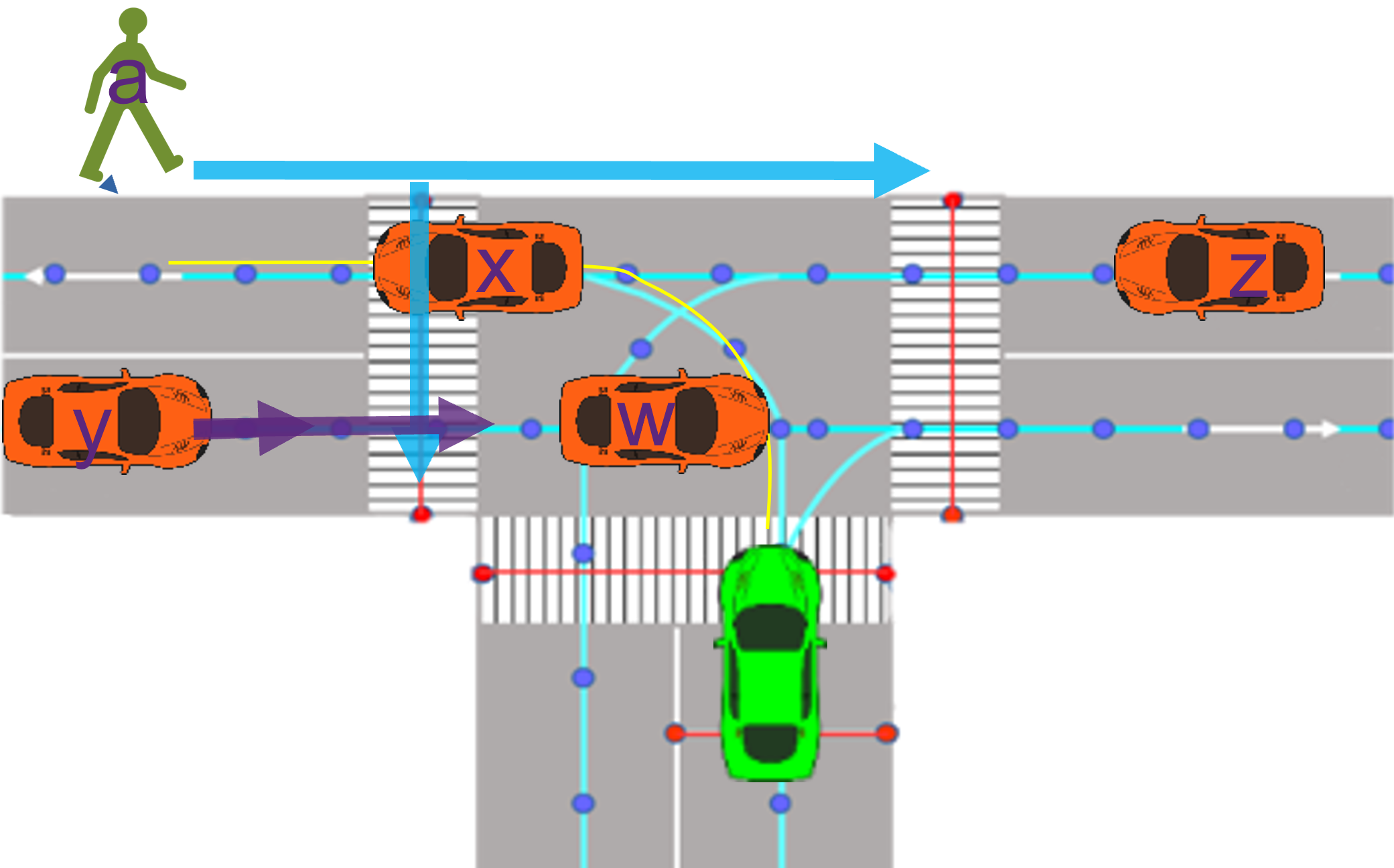}
\caption{Traffic Scene for our example.}
\label{fig:example_scene}
\end{figure}

\begin{figure} [ht]
\centering
\includegraphics[width=01.0\columnwidth]{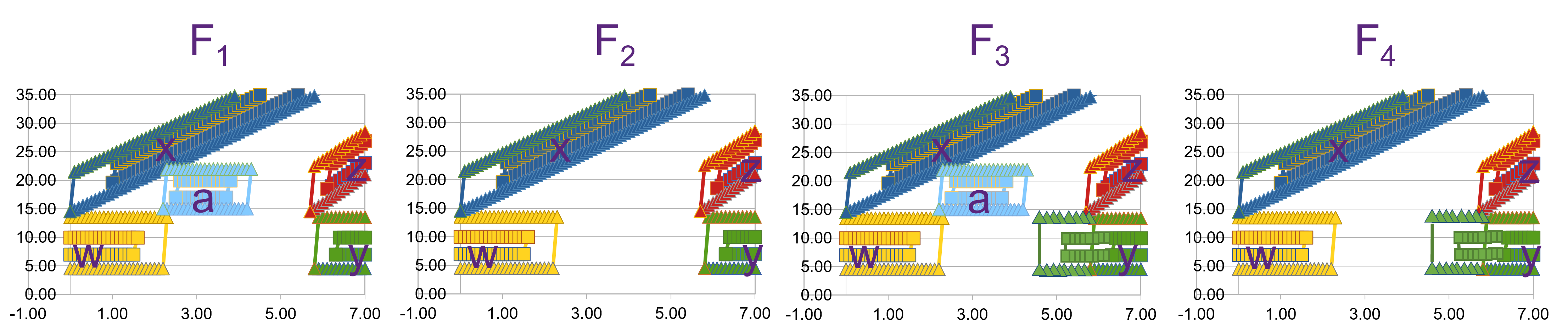}
\caption{The four futures, displayed as ST graphs, corresponding to our traffic scene example.}
\label{fig:example_futures}
\end{figure}

\begin{figure} [ht]
\centering
\includegraphics[width=01.0\columnwidth]{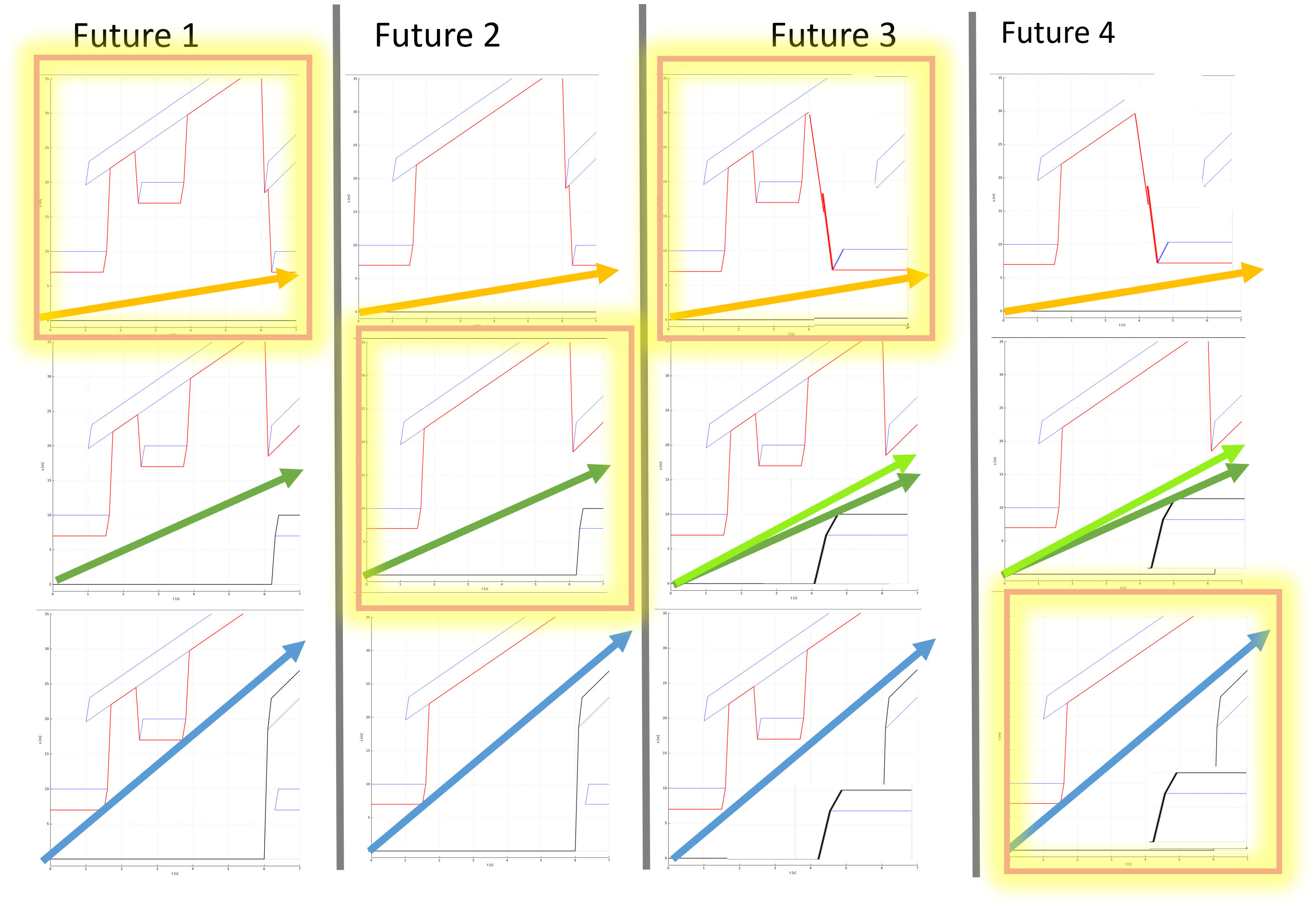}
\caption{The different convex bounds corresponding to each future. The orange highlight shows one possible selection of bounds that could be used for an instance of multi-future MPC.}
\label{fig:example_bounds}
\end{figure}

\begin{figure} [ht]
\centering
\includegraphics[width=01.0\columnwidth]{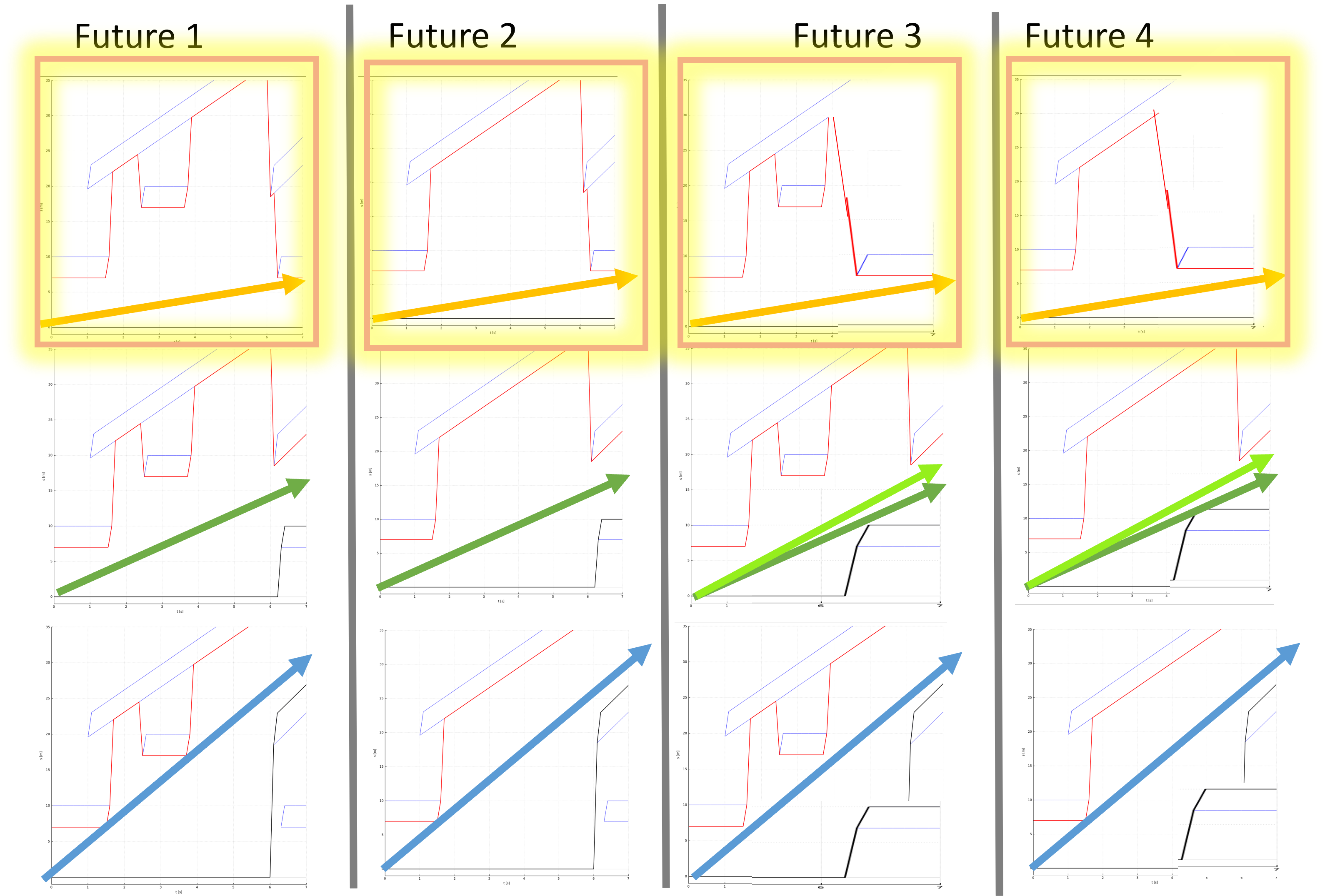}
\caption{A depiction of pairing closest basins.}
\label{fig:example_pairing}
\end{figure}

In our scene in Fig.~\ref{fig:example_scene}, there are $k=5$ traffic agents. The number of futures is $m=4$ and is shown in Fig.~\ref{fig:example_futures}. Futures could be provided as joint predictions from our prediction algorithm, but in the case that we use a prediction algorithm that gives single agent predictions, $O(m)=r^k$ where $r$ is the number of single agent predictions. In our example the pedestrian $a$ and car $y$ have two predictions each, and the other cars have single predictions, not visualized for clarity. 
The constraints corresponding to the four futures are not convex. There can be $O(2^k)$ constraints for each future. This upper bound is $2^5=32$ with five agents, however in practice, this is often much lower based on the ordering of which agents cross our path and when. Figure \ref{fig:example_bounds} shows that in our example, there are three convex bounds $\bar{c}$ for each future.   
When solving for multiple futures in parallel, we take as input a bound for each future. However since there are multiple bounds for each future, we should consider every possible combination of bounds, resulting in $O(\bar{c}^m)$ multi-future problems. An arbitrary combination of bounds is highlighted in orange in Fig.~\ref{fig:example_bounds}. To reduce this to $O(\bar{c})$, we pair the closest basins for each problem as shown in Fig.~\ref{fig:example_pairing}. The colored arrows indicate the basins that will be paired. An example pairing is highlighted in orange. Note from the two green arrows in future 3 and 4 that the basins may not exactly match. To determine the closest basins, we use the euclidean distance on the approximate solutions generated by Algorithm \ref{alg:approx}.
In our example, this reduces the $3^4=81$ multi-future optimizations to $3$, however note that this is a heuristic approximation, and is no longer guaranteed to be optimal.

\subsection{Greedy decision making is suboptimal}
\begin{figure}[ht] 
\centering
\includegraphics[width=\columnwidth]{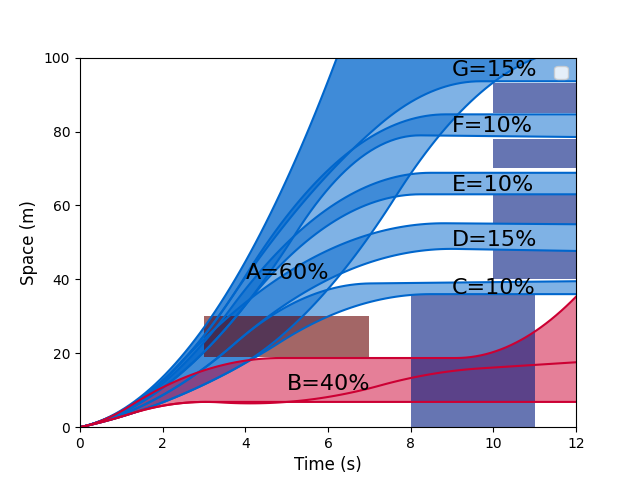}
\caption{Making decisions based on probabilities at the first decision point can be greedy. For example, while future A is more probable, selecting future A possibly forces us to other choices that have a comparatively lower likelihood.}\label{fig:greedy}
\end{figure}

If a scene has multiple decision points, selecting the most probable future at the first decision point is potentially greedy, as depicted in the Fig.~\ref{fig:greedy}

\subsection{Discussion on interactivity and game theory}
While in this work, we assume information gain is only influenced by time, indicating our preferred action more strongly could be used as communication tool to not only probe but also influence other traffic agents. For example, in our formulation, the first $t_d$ time steps are locked and biased by the probabilities of each future. Because we are trying to keep all options open, this could make the resulting behavior more difficult to interpret for other traffic participants. 

To increase interpretability, there is a potential to overweight a preferred trajectory. Conversely, if we are in an adversarial setting, we might wish to obfuscate or even deceive others concerning our intended action. An equal weighting could obfuscate an agent's intention, while strongly weighting a less likely intention could deceive or otherwise influence other agents. 
While we don't explore these ideas in the current work, we flag them as a potential target for future research. 

We believe that the probabilities of different futures can be conditioned on our action. In this way, we could expand our existing model to reason about probing actions for interactive decision making. 

\subsection{Limitations}
In this work, we do not consider the interactivity of driving. In driving, sometimes we deliberately make our intention clear. Selecting a trajectory that keeps the ego agent as flexible as possible can make it difficult for other agents to predict our action. We leave it as future work to explore how we can clear;y indicate our intention while remaining flexible.






\end{document}